\documentclass[11p, table]{article}
\usepackage[preprint]{coling}

\usepackage{booktabs}
\usepackage{blindtext}
\usepackage{placeins}
\usepackage{times}
\usepackage{latexsym}
\usepackage{todonotes}
\usepackage{float}
\usepackage{subfig}
\usepackage{arydshln}
\usepackage{tabularx}
\usepackage{xcolor}

\usepackage[T1]{fontenc}

\usepackage[utf8]{inputenc}

\usepackage{microtype}

\usepackage{inconsolata}

\usepackage{graphicx}

\title{Language verY Rare for All}

\author{
  \textbf{Ibrahim Merad\textsuperscript{1}},
  \textbf{Amos Wolf\textsuperscript{2}},
  \textbf{Ziad Mazzawi\textsuperscript{2}},
  \textbf{Yannick Léo\textsuperscript{1,2}}
\\
\\
  \textsuperscript{1}Kaukana Ventures,
  \textsuperscript{2}Emerton Data,
\\
}

\begin{document}
\maketitle
\begin{abstract}
In the quest to overcome language barriers, encoder-decoder models like NLLB have expanded machine translation to rare languages, with some models (e.g., NLLB 1.3B) even trainable on a single GPU. While general-purpose LLMs perform well in translation, open LLMs prove highly competitive when fine-tuned for specific tasks involving unknown corpora. We introduce LYRA (Language verY Rare for All), a novel approach that combines open LLM fine-tuning, retrieval-augmented generation (RAG), and transfer learning from related high-resource languages. This study is exclusively focused on single-GPU training to facilitate ease of adoption. Our study focuses on two-way translation between French and Monégasque — a rare language unsupported by existing translation tools due to limited corpus availability. Our results demonstrate LYRA’s effectiveness, frequently surpassing and consistently matching state-of-the-art encoder-decoder models in rare language translation.
\end{abstract}

\section{Introduction}
\label{sec:intro}
Machine translation has come a long way since its inception in the 1940s. The methodology evolved from the initial rule-based approach~\cite{hutchins1986machine, hutchins1997first} to statistical machine translation~\cite{brown1993mathematics, koehn2009statistical} and most recently adopted neural systems as the de-facto approach yielding superior results~\cite{bahdanau2014neural, cho2014properties}. An important breakthrough occurred with the advent of Transformers~\cite{vaswani2017attention} whose attention-based architecture did not only allow for better translation but paved the way for an NLP revolution through LLMs~\cite{brown2020language, radford2018improving, minaee2024large}. The considerable progress observed on a wide range of NLP tasks is the combined result of the ingenuous Transformer neural architecture, the availability of large GPU compute resources and macroscopic amounts of training data. However, the uneven data amounts between different languages translate to varying performances on NLP tasks~\cite{joshi-etal-2020-state, blasi-etal-2022-systematic}, including machine translation. Thus, contrary to widespread languages for which large text corpora are available including parallel data, lesser known languages suffer from data scarcity which makes it difficult to train deep learning models~\cite{zhang2020neural}. Moreover, compensating this inequality by obtaining data for low resource languages is expensive and logistically challenging~\cite{nekoto-etal-2020-participatory, kuwanto2023low, orife2020masakhane}.

This work is concerned with training a neural machine translator between the French and Monégasque language. A very low resource language only spoken by around 5,000 people to date in the Principality of Monaco and which, to our knowledge, remains uncovered by any neural machine translator. We take on the task of creating a parallel French-Monégasque dataset enabling the training of translators and language models on this language. 
We finetune multiple models on this task and present our methodology called LYRA allowing to optimize results with limited data (about 10K parallel sentences and a dictionary).

\begin{figure*}[h]
    \centering
    \includegraphics[width=\linewidth]{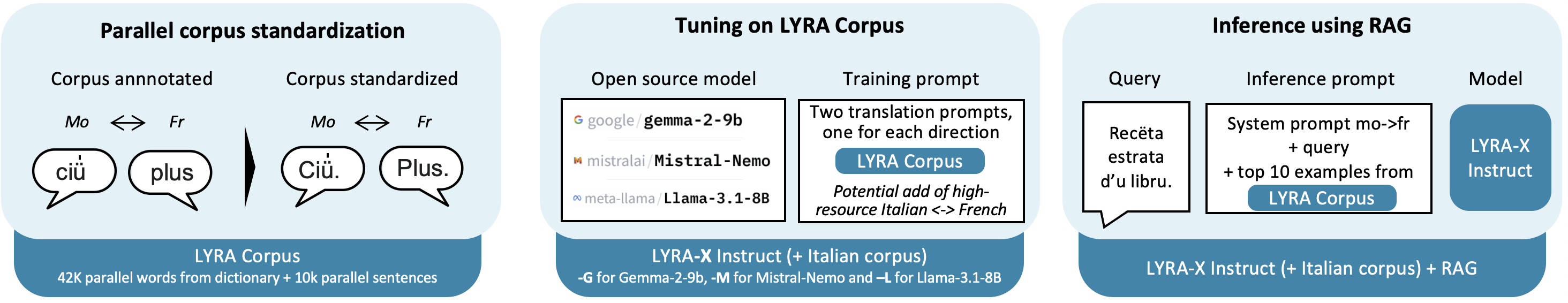}
    \caption{Illustration of our method for building LYRA.}
    \label{fig:method}
\end{figure*}

\section{Related works}
\label{sec:related}

Given the challenge it poses, the low-resource setting has received much attention in the literature~\cite{haddow2022survey, hedderich2020survey, magueresse2020low}. 
The proposed strategies include targeted data gathering~\cite{hasan2020not}, exploiting monolingual data~\cite{gibadullin2019survey}, backtranslation~\cite{sennrich2015improving}, transfer learning~\cite{dabre2020survey, zoph2016transfer} and multilingual models~\cite{johnson2017google}.

The most notable effort towards a model with high language coverage is NLLB~\cite{costa2022no} (No Language Left Behind). 
The latter translator was trained for pairs among over 200 different languages using a Sparsely Gated Mixture of Experts architecture. For this purpose, the authors created the Flores-200 dataset consisting of 3000 parallel sentences establishing a benchmark for multilingual machine translation. However, this effort did not include the Monégasque language. 

While NLLB uses an encoder-decoder architecture specifically intended for translation, decoder-only models also reached competitive performance on multiple tasks including translation~\cite{hendy2023good, wei2022emergent, NEURIPS2022_b1efde53}. This motivated works to improve results with such models~\cite{xu2024a, yang2023bigtranslate, alves2024tower} since they offer a far more interesting option due to their higher flexibility and impressive multitasking abilities~\cite{10.1145/3411763.3451760, NEURIPS2022_8bb0d291, NEURIPS2021_5c049256}. Moreover, decoder-only models can leverage strategies like RAG to improve performance and enjoy greater attention in the literature leading to faster progress. Finally, these models hold the same potential for multilingual translation and transfer learning. Nonetheless, these references did not consider low-resource languages.
 

Most recently, both model types were combined by GenTranslate~\cite{hu2024gentranslate} which uses a Seq2Seq model to sample translations that are fed into an LLM to combine them into an improved answer. Note however that this work assumes high compute resources with multiple GPUs.

In this work, NLLB as well as a few open LLMs are finetuned using LYRA on a newly created French-Monégasque dataset using only a single GPU machine. We compare their performances on this translation task and showcase the benefits of LYRA in the low-resource setting.

\section{Data}\label{sec:data}

Since we are unaware of any preexisting parallel corpus involving Monégasque, we created a French-Monégasque dataset using OCR from a few sources including: A French-Monégasque dictionary, a Monégasque grammar book, as well as a few literary works available in both languages. These include works such as the poem collection ``Lettres de mon moulin'', the play ``Antigone'' and some Tintin comics. The acquired inputs were later combined into parallel entries via manual annotation.

The dataset contains a total number of 10,794 parallel French-Monégasque sentences in addition to 42,698 entries from the dictionary and the grammar book which includes verb conjugations and proverbs. The test set was constituted by selecting sentences with high quality translation in order to ensure a reliable basis for evaluation.

The fact that this unique existing dataset has under 100K pairs makes the Monégasque language a very low resource language based on the conventions adopted by~\citet{costa2022no}.

\section{Methods}
\label{sec:methods}
The LYRA methodology, illustrated on Figure~\ref{fig:method}, aims to maximize translation quality in the low data context using three main strategies.

\paragraph{Leveraging related high-resource languages} Previous works demonstrated the benefits of knowledge transfer in multilingual neural machine translation~\cite{dabre2020survey, zoph2016transfer, maimaiti2019multi}. In order to take advantage of this phenomenon, we perform a preliminary finetuning phase on translation between French and Italian, which is a high resource language pair, before finetuning on French-Monégasque translation. The idea is to exploit the grammatical similarity between Monégasque and Italian. Thus, in the preliminary phase, the model learns to transition between French and Italian-like grammatical structures on plentiful data which facilitates the subsequent finetuning on French-Monégasque translation. 


\begin{figure*}[t]
    \centering
    \includegraphics[width=0.8\linewidth]{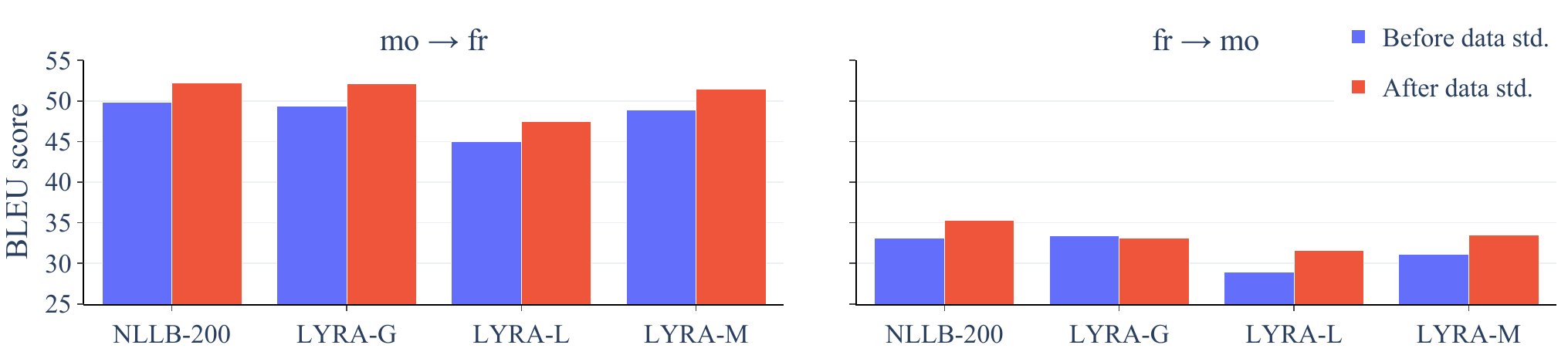}
    \caption{Comparison of models' translation performance in both directions in terms of BLEU scores before and after data standardization. The latter uniformly improves translation performance across all models.}
    \label{fig:st_vs_prest}
\end{figure*}

\paragraph{Data standardization}

As often emphasized, training models for NLP applications considerably depends on data quality to achieve high performance~\cite{tokpanov2024zyda, hoffmann2022training, rae2021scaling}. This aspect is all the more important when data is scarce. We measure the impact of careful data curation in the current setting by training the candidate models on two versions of the French-Monégasque dataset. The initial raw version featured some issues of inconsistent capitalization and punctuation and used various quotation marks. 
The impact of these details on downstream performance should not be underestimated since they can confuse the model by causing irregular tokenization.


Considering the potential performance gain, we invest the effort of standardizing the sentences in the first version of the dataset to fix these issues and obtain a curated second version.

\paragraph{Retrieval Augmented Generation} For decoder-only models, the training data can be used to improve test-time performance by including the most similar sentence or word pairs into the prompt. Note that this is akin to few-shot prompting but using embeddings to retrieve the most similar examples. 
Since the Monégasque language is unknown to the available embedding models, the French parts are used to generate an embedding for each instance. For this purpose, words and sentences are encoded using a high performing model on French retrieval tasks. The latter is available on the HuggingFace Hub under the reference \texttt{BAAI/bge-multilingual-gemma2}. Retrieval of the nearest neighbors is then carried out based on cosine similarity. The number of retrievals is fixed to 10 instances.


\section{Experiments}
\label{sec:exp}

The impact of each strategy on translation quality is evaluated by testing them sequentially. The effect of data standardization is measured prior to testing the other strategies. Performance is measured using the BLEU score~\cite{papineni2002bleu} as well as METEOR which is more correlated with human assessment~\cite{banerjee2005meteor}. We also provide evaluations using the chrF++ metric~\cite{popovic2015chrf} in Appendix~\ref{sec:additional_results}.

\paragraph{Models} The focus is set on single-GPU training to make the experiments more relevant for the low resource context. We finetune some high-performing models on French-Monégasque translation and assess the performance gains from each strategy. The distilled model \texttt{nllb-200-distilled-1.3B} was chosen as a representative of the NLLB encoder-decoder model family since it outperforms the 3B model and reaches close performance to the original 54B model at much lower computational costs~\cite{costa2022no}. As for decoder-only models, the candidates are the public LLMs : \texttt{Llama-3.1-8B}~\cite{dubey2024llama} (LYRA-L), \texttt{gemma-2-9b} (LYRA-G) and \texttt{Mistral-Nemo-Instruct-2407} 12B (LYRA-M). This choice targets high performing models which have benefited from multilingual pretraining, including French and Italian (to which Monégasque is related), while keeping our compute budget in mind. The LLMs are finetuned using LoRA~\cite{hu2021lora} with the efficient implementation of the \texttt{unsloth} library.

Given that Monégasque was not among the languages covered by NLLB, the \texttt{nllb-200-distilled-1.3B} model is finetuned using the French-Monégasque data. In order to maximize downstream performance, we use NLLB's Ligurian tokenizer on Monégasque sentences. The rationale behind this choice is that Ligurian (another low resource language related to genoese) is an even closer language to Monégasque than Italian. Therefore, using the Ligurian tokenizer is likely to yield a more useful representation of Monégasque text.
All the presented experiments use greedy decoding.


\paragraph{Effect of Data standardization} The candidate models are trained on both versions of the French-Monégasque dataset and evaluated on translation in both directions. 
Figure~\ref{fig:st_vs_prest} compares the performances reached by each model by training on the dataset before and after undergoing standardization. We observe that all models improve their scores by a significant amount thanks to the standardized data.

We also note that translation quality is clearly superior towards the French language. This is explained by the fact that most models were pretrained on plentiful amounts of French text allowing them to master this high-resource language beforehand. On the other hand, they only discover the Monégasque language through our small dataset which limits the proficiency they are able to reach.

\paragraph{Effect of RAG} As previously mentioned, the \texttt{BAAI/bge-multilingual-gemma2} model is used in order to generate embeddings of the French sentences. This is done for the train and test sets and the embeddings are used to improve test-time performance by retrieving, for each test sample, the 10 nearest train samples and including them in the prompt. Obviously, this can only be done for LLMs and not for NLLB. The models are trained on the standardized data and their BLEU and METEOR scores with and without RAG are reported on Table~\ref{tab:all}.

Significant improvements in BLEU scores are seen for translation towards French across all models after the addition of RAG. However, LYRA-G is the only one to benefit from RAG for the fr$\to$mo direction while LYRA-M suffers a significant degradation of its score. These observations may be explained by the fact that the embeddings are based on the French part of the data only and that the embedding model is originally based on Gemma 2.
\begin{table}[h]
    \fontsize{30pt}{40pt}\selectfont
  \centering
  \resizebox{1.02\columnwidth}{!}{
  \begin{tabular}{|l|cc|cc|cc|cc}
    \hline
    \textbf{Model} & \multicolumn{2}{c|}{BLEU} & \multicolumn{2}{c|}{METEOR} \\
     & fr→mo & mo→fr & fr→mo & mo→fr\\
    \hline
    NLLB-200 1.3B & \textbf{35.27} & 52.18 & 48.17 & 63.55 \\
    \hline
    LYRA-L Instruct & 31.62 & 47.49 & 49.35 & 65.20 \\
    \hspace{10pt}+ RAG & 31.32 & \underline{52.67} & 49.45 & \underline{70.04} \\
    \hspace{10pt}++ Italian corpus & \underline{32.83} & 51.95 & \underline{50.79} & 69.07 \\
    \hline
    LYRA-G Instruct & 33.16 & 52.12 & 51.47 & 69.40 \\
    \hspace{10pt}+ RAG & 34.42 & \textbf{\underline{58.10}} & 52.91 & \textbf{\underline{74.31}} \\
    \hspace{10pt}++ Italian corpus & \underline{35.25} & 57.23 & \textbf{\underline{53.19}} & 73.36 \\
    \hline
    LYRA-M Instruct & \underline{33.46} & 51.49 & \underline{51.77} & 69.02 \\
    \hspace{10pt}+ RAG & 30.69 & \underline{56.75} & 48.38 & \underline{72.38} \\
    \hspace{10pt}++ Italian corpus & 32.31 & 54.88 & 49.31 & 70.97 \\
    \hline
  \end{tabular}
  }
  \caption{Translation performance in both directions as measured by BLEU and METEOR scores using the standardized data and other methods. Bold numbers represent best scores among all models.
  }
  \label{tab:all}
\end{table}

\paragraph{Effect of French-Italian finetuning} We finally evaluate the effect of a preliminary finetuning phase on French-Italian translation before training on the French-Monégasque data. This recipe is tested using the \texttt{opus-books} dataset~\cite{4992de1b5fb34f3e9691772606b36edf} which contains high quality French-Italian parallel sentences. NLLB is excluded from this experiment since it is considered to have already benefited from transfer learning. Indeed, NLLB was pretrained on over 200 languages including French, Italian and Ligurian which is even closer to Monégasque. 

The scores of models trained in this fashion and tested with RAG are reported on Table~\ref{tab:all} (omitting RAG led to inferior results). A clear benefit is observed on fr$\to$mo scores for LYRA-L and LYRA-G which lets the latter virtually match NLLB's BLEU score. However, LYRA-M still attains its best fr$\to$mo score in the base setting. On the other hand, some performance is lost in the mo$\to$fr direction. We posit that the LLMs' pretrained proficiency in French slightly degrades after undergoing a finetuning procedure involving two other languages.

\section{Conclusion}

In this work, we presented LYRA, a methodology to boost machine translation performance despite scarce data. 
We saw that enhancing data quality effectively improved results in general. RAG also showed significant potential although some model specific adaptation may sometimes be necessary. Finally, we have also seen that models can reach higher proficiency in a low resource language thanks to transfer learning. Further gains will likely be possible by finetuning future higher performing LLMs.
Finally, data augmentation is another interesting research avenue to deal with the low-resource setting.

\section{Limitations}
Although the results confirm the benefits of the presented methodology, the latter still has its limitations. For example, data curation cannot improve performance beyond a certain point and should be combined with data augmentation to alleviate data scarcity. Moreover, RAG can only help performance if train data are diverse enough and include relevant examples. Finally, not all low-resource languages are related to high resource ones so that transfer learning will not always be useful.


\section*{Acknowledgments}
We would like to extend our special thanks to the annotation teams from Afuté and Isahit for their hard work, the Monégasque experts from the Comité des Traditions, the Government of Monaco—particularly the Délégation Interministérielle chargée de la Transition Numérique—and the FAIR team at Meta, including Alexandre Mourachko, for their invaluable advice on the NLLB project.

\bibliography{custom}

\appendix

\section{Additional results}\label{sec:additional_results}

The performances of the trained models as measured by the chrF++ metric~\cite{popovic2015chrf} are reported on Table~\ref{tab:chrF++}. These figures mostly agree with BLEU scores when comparing the models.

\begin{table}[h]
  \centering
  \begin{tabular}{|l|cc|cc}
    \hline
    \textbf{Model} & \multicolumn{2}{c|}{chrF++} \\
     & fr→mo & mo→fr \\
    \hline
    NLLB-200 1.3B before std. & 55.61 & 65.59 \\
    NLLB-200 1.3B & \textbf{\underline{57.90}} & \underline{67.05} \\
    \hline
    LYRA-L before std. & 50.87 & 61.47 \\
    LYRA-L & 53.26 & 63.90 \\
    \hspace{10pt}+ RAG & 53.78 & \underline{68.03} \\
    \hspace{10pt}++ Italian corpus & \underline{54.81} & 66.99 \\
    \hline
    LYRA-G before std. & 55.32 & 66.51 \\
    LYRA-G & 55.48 & 67.87 \\
    \hspace{10pt}+ RAG & \underline{57.32} & \textbf{\underline{71.89}} \\
    \hspace{10pt}++ Italian corpus & 57.16 & 71.55 \\
    \hline
    LYRA-M before std. & 53.63 & 65.19 \\
    LYRA-M & \underline{55.44} & 67.55 \\
    \hspace{10pt}+ RAG & 52.11 & \underline{69.75} \\
    \hspace{10pt}++ Italian corpus & 54.02 & 69.42 \\
    \hline
  \end{tabular}
  \caption{Translation performance in both directions as measured by chrF++ scores using the standardized data and other methods. Bold numbers represent best scores among all models. After preliminary finetuning on French-Italian data, all models achieved superior results using RAG rather than without.}
  \label{tab:chrF++}
\end{table}

\begin{figure*}[h]
    \centering
    \includegraphics[width=\linewidth]{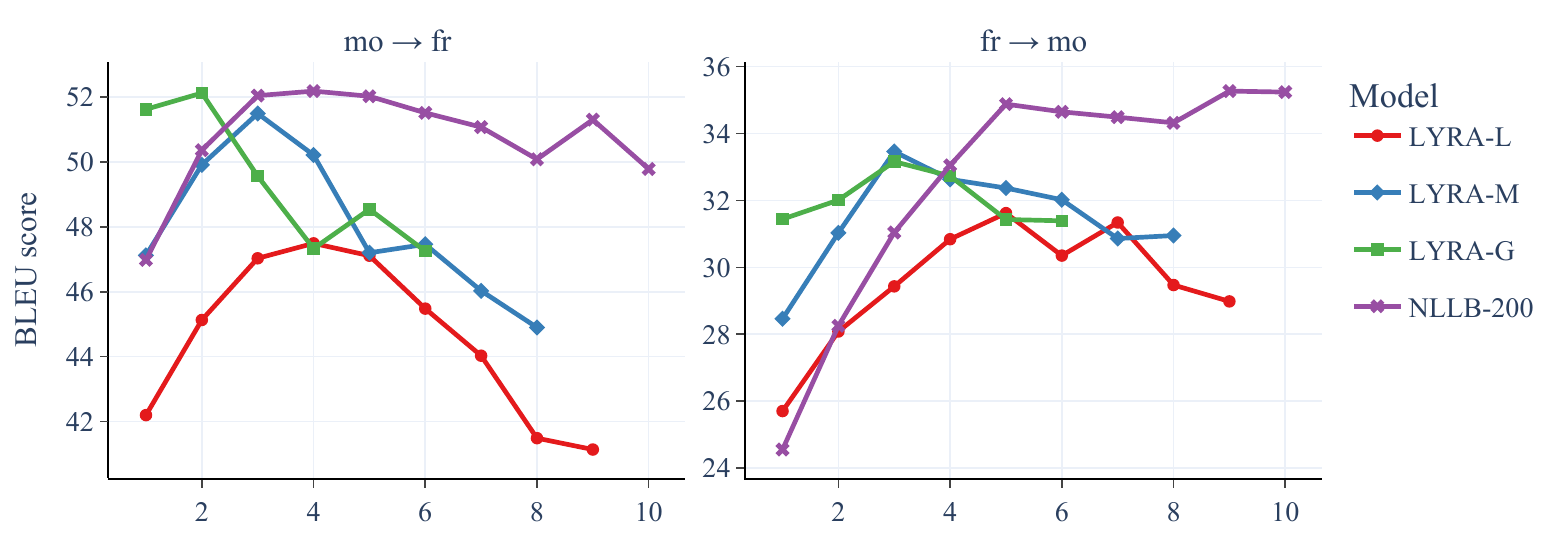}
    \caption{Evolution of translation performance in both directions for the considered models through training epochs as measured by the BLEU score. The training of certain models was stopped early due to overfitting.}
    \label{fig:bleu_evolution}
\end{figure*}

Figure~\ref{fig:bleu_evolution} displays the evolution of BLEU scores on translation in both directions through training epochs. One can observe that, apart from NLLB, most models quickly overfit the data due to their limited quantity. 

\section{Additional data details}
We provide below a list of the sources used to constitute the French-Monégasque parallel dataset on which the models were trained:
\begin{itemize}
    \item A French-Monégasque dictionary containing two-way translations of single words as well as proverbs.
    \item A Monégasque grammar book (Monégasque Bescherelle) containing verb conjugations and their translations into French.
    \item The ``Üntra Nui'' stories which is a Monégasque chronicle.
    \item Poems \& Fables from Monégasque culture.
    \item The play ``Antigone'' written by Jean Anouilh.
    \item The collection of short stories titled ``Lettres de mon moulin'' by Alphonse Daudet.
    \item A collection of Monégasque songs.
    \item 3 chapters of Tintin comics available in both languages. Namely :
    \begin{itemize}
        \item ``Le secret de la Licorne''.
        \item ``Le trésor de Rackham le Rouge''.
        \item ``Les bijoux de la Castafiore''.
    \end{itemize}
\end{itemize}

\begin{table*}[t!]
\begin{tabularx}{\linewidth}{|X|X|}
   \hline
    \multicolumn{1}{|c|}{Monégasque} & \multicolumn{1}{c|}{French} \\
   \hline
    \cellcolor{red!10}Ah!... M' asperavi?... Savi dunca perche sun aiçi ?..  &  \cellcolor{red!10}Ah?... Vous m'attendiez? Vous connaissez donc le but de ma visite?.. \\
   \cdashline{1-2}
   \cellcolor{green!10}Ah ! M' asperavi ? Savi dunca perche sun aiçi ?  & \cellcolor{green!10}Ah ? Vous m'attendiez ? Vous connaissez donc le but de ma visite ? \\
   \hline
   \cellcolor{red!10}A grafia e tamben ë tradüçiue d’i testi d’achëstu calendari sun de l’autu sarvu a tradüçiun d’u puema «O belu Munegu» 	& \cellcolor{red!10}La graphie ainsi que les traductions des textes de ce calendrier sont de l’auteur excepté la traduction du poème «Ô Monaco la belle» 	\\
   \cdashline{1-2}
    \cellcolor{green!10}A grafia e tamben ë tradüçiue d’i testi d’achëstu calendari sun de l’autu sarvu a tradüçiun d’u puema "O belu Munegu". 	& \cellcolor{green!10}La graphie ainsi que les traductions des textes de ce calendrier sont de l’auteur excepté la traduction du poème "Ô Monaco la belle".\\
   \hline
   \cellcolor{red!10}Ancœi, a Cumpagnia e cumpusa de trei ufiçiali, düjanœve suta-ufiçiali e nuranta sete surdati & \cellcolor{red!10}Actuellement son effectif est de trois officiers, 19 sous-officiers et 97 hommes du rang 	\\
   \cdashline{1-2}
    \cellcolor{green!10}Ancœi, a Cumpagnia e cumpusa de trei ufiçiali, düjanœve suta-ufiçiali e nuranta sete surdati. & \cellcolor{green!10}Actuellement son effectif est de trois officiers, dix-neuf sous-officiers et quatre vingt dix-sept hommes du rang.\\
    \hline
    \cellcolor{red!10}E a fau tanta paciara, De « ci, ci », e de ci, cia » Ch’ün caciaire, che passava Gh’a futüu üna füsiya ! 	& \cellcolor{red!10}Et il fit tellement de potin, Des « ci, ci » et des « ci, cia », Qu’un chasseur qui passait, L’abattit d’un coup de fusil. 	\\
   \cdashline{1-2}
    \cellcolor{green!10}E a fau tanta paciara, ch’ün caciaire, che passava gh’a futüu üna füsiya. & \cellcolor{green!10}Et il fit tellement de potin qu’un chasseur qui passait, l’abattit d’un coup de fusil.\\
   \hline
\end{tabularx}
  \caption{Example instances from the French-Monégasque dataset before (red cells) and after standardization (green cells).}
  \label{tab:std}
\end{table*}
Table~\ref{tab:std} shows a few examples of sentence pairs before and after undergoing standardization. These illustrate the fixed issues including excessive use of ellipsis, non standard quotes, digital instead of literal numbers and arbitrary onomatopoeia.

The full dataset (before and after standardization) can be found in the following github repository \url{https://github.com/EmertonData/lyra}.
\section{Experimental details}

All the models were trained using a single Nvidia A100 40 GB GPU. NLLB-200 1.3B was finetuned with learning rate: $10^{-5}$ and batch size 32.

Regarding the LLMs, the 4 bit quantized versions provided by \texttt{unsloth} were used as starting points and finetuned with this library using LoRA with the following configuration:
\begin{itemize}
    \setlength\itemsep{-4pt}
    \item \texttt{r = 16}
    \item \texttt{lora\_alpha = 16}
    \item \texttt{lora\_dropout = 0.0}
    \item \texttt{bias = "none"}    
    \item \texttt{target\_modules = ["q\_proj", "k\_proj", "v\_proj", "o\_proj", "gate\_proj", "up\_proj", "down\_proj"]}
    \item \texttt{use\_rslora = True}
    \item \texttt{loftq\_config = None} 
\end{itemize}

A learning rate equal to 3e-5 was used for LYRA-G and 1e-5 for LYRA-L and LYRA-M. Apart from that, the following training parameters are common:
\begin{itemize}
    \setlength\itemsep{-4pt}
    \item \texttt{batch\_size = 48}
    \item \texttt{packing = False}
    \item \texttt{warmup\_steps = 100}
    \item \texttt{optim = "adamw\_8bit"}
    \item \texttt{weight\_decay = 0.01}
    \item \texttt{lr\_scheduler\_type = "cosine"}
    \item \texttt{max\_seq\_length = 2048}
\end{itemize}

All LLMs were trained on completions only using the appropriate data collator. Training was launched for 10 epochs but early stopping was performed based on validation loss as seen on Figure~\ref{fig:bleu_evolution}.

\end{document}